\documentclass[10pt,twocolumn,letterpaper]{article}

\usepackage{cvpr}
\usepackage{times}
\usepackage{epsfig}
\usepackage{graphicx}
\usepackage{amsmath}
\usepackage{amssymb}

\usepackage[breaklinks=true,bookmarks=false]{hyperref}

\cvprfinalcopy 

\def\cvprPaperID{3997} 

\ifcvprfinal\fi
\begin{document}

\title{Grid R-CNN Plus: Faster and Better}

\author{Xin Lu$^1$ \quad Buyu Li$^2$ \quad Yuxin Yue$^3$ \quad Quanquan Li$^1$ \quad Junjie Yan$^1$\\
$^1$SenseTime Research,  $^2$The Chinese University of Hong Kong,~$^3$Beihang University\\
{\tt\small \{luxin,liquanquan,yanjunjie\}@sensetime.com,  byli@ee.cuhk.edu.hk,  yueyuxin@buaa.edu.cn}
}

\maketitle
\thispagestyle{empty}

\begin{abstract}
Grid R-CNN is a well-performed objection detection framework. It transforms the traditional box offset regression problem into a grid point estimation problem. With the guidance of the grid points, it can obtain high-quality localization results. However, the speed of Grid R-CNN is not so satisfactory.
In this technical report we present Grid R-CNN Plus, a better and faster version of Grid R-CNN. We have made several updates that significantly speed up the framework and simultaneously improve the accuracy.
On COCO dataset, the Res50-FPN based Grid R-CNN Plus detector achieves an mAP of 40.4\%, outperforming the baseline on the same model by \textbf{3.0} points with similar inference time. Code is available at \href{https://github.com/STVIR/Grid-R-CNN}{https://github.com/STVIR/Grid-R-CNN}.

\end{abstract}

\section{Introduction}
\label{sec:intro}

Grid R-CNN~\cite{Lu_2019_CVPR} achieves state-of-the-art performance on the object detection task. Traditional object detectors treat the localization task as a bounding box offset regression problem \cite{girshick2015fast,ren2015faster,lin2017feature,he2017mask}. In contrast, Grid R-CNN utilizes explicit spatial representation to estimate the location of the grid points, thus reformulating the localization as a classification problem. And the grid points guided localization mechanism enables it to perform surprisingly well in the high-quality detection tasks.
As described in ~\cite{Lu_2019_CVPR}, Grid R-CNN significantly improve the accuracy of localization at strict localization criteria (IoU with ground truth higher than 0.75). Note that the region proposal network and classification branch are  maintained, so that the improvement just comes from the new designed localization mechanism.

Although Grid R-CNN is excellent in accuracy, the non-negligible computation cost prevents it to become a widely used framework. Thus we have made further study to improve it. In this technical report, we propose Grid R-CNN Plus, which integrates several effective changes that lead to much higher speed as well as increase in detection accuracy. 

Grid R-CNN uses the same representation region to generate the supervision maps for all the grid-points. But it is inefficient. For example, the top-left grid point will not nearly appear on the right or bottom side of the supervision map. Thus in Grid R-CNN Plus we introduce the grid point specific representation region. Only the 1/4 most probable region is used for supervision. With the representation region reduced, we hence reduce the size of feature maps in the grid branch. Moreover, the number of convolution layers for the grid feature fusion is also reduced. So the computation cost is further reduced. Simultaneously the accuracy of grid location is increased owing to the representation concentration.

We have also made further analysis and modifications on sampling strategy, normalization method, NMS strategy and some hyper-parameters. The goal of all the updates is to make model faster and better. And Grid R-CNN Plus achieves similar inference speed to Faster R-CNN FPN, with significant gain in accuracy.

\begin{figure}[ht]
\centering
\includegraphics[width=1\linewidth]{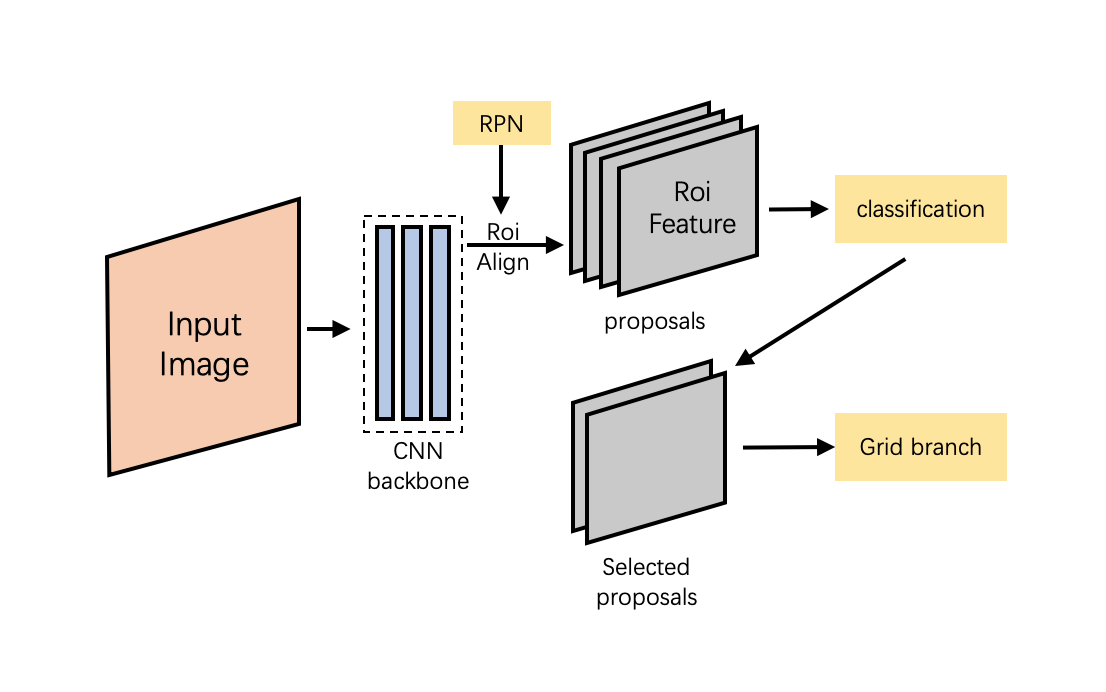} 
\caption{\textbf{Overview of the pipeline of Grid R-CNN.} Region proposals are obtained from RPN and used for RoI feature extraction from the output feature maps of a CNN backbone. The RoI features are then used to perform classification and localization. Like mask-rcnn~\cite{he2017mask}, proposals are first filtered by non-maximum suppression by classification score. Then top K proposals will be sent to grid branch for high quality localization.}
\label{fig:pipeline}
\end{figure}

\section{Revisiting Grid R-CNN}
Figure~\ref{fig:pipeline} shows the overview of Grid R-CNN framework. Like other two-stage detectors, Grid R-CNN consists of RPN (region proposal network) and R-CNN. Based on region proposals, features for each RoI are extracted individually by RoIAlign operation from the output of the CNN backbone (e.g., ResNet~\cite{he2016deep}). These RoI features are then used to predict the classes and locations (usually the offsets) of the corresponding proposals. In contrast to most previous works, e.g., Faster R-CNN~\cite{ren2015faster}, we locate objects by predicting grid points instead of bounding box offset. The grid prediction branch adopts a fully convolutional neural  network~\cite{long2015fully}. It outputs a fine-grained spatial layout (probability heatmap) from which we can locate the grid points of the object. With the grid points, we can therefore determine the accurate object bounding box.

It is worth noting that the grid branch adopts several consecutive convolutions and deconvolutions to generate the final grid points heatmaps. So this branch could be heavy. Previous works like Mask R-CNN~\cite{he2017mask} adopt an additional sampling approach to reduce computation on heavy branches (e.g., mask branch and key-point branch). Grid R-CNN uses a similar strategy: first the 1000 proposals (selected from RPN results) are forwarded to the classification branch to obtain the scores, and then NMS is performed and the top 100 of the left proposals are forwarded to the grid branch. Thus the grid branch becomes not so time consuming.

To improve the performance, grid points feature fusion and extended region mapping mechanism are proposed in Grid R-CNN. Grid points feature fusion module takes advantage of spatial correlation between grid points to calibrate "hard" grid points. Extended region mapping mechanism enables the grid branch to handle the case where ground truth points lie outside of the RoI. It redefines the representation area of the output heatmap as a four times larger corresponding region in the image, so that all grid points are covered in most cases. More details can be found in ~\cite{Lu_2019_CVPR}.

\section{Grid R-CNN Plus}
\begin{figure}[ht]
\centering
\includegraphics[width=1\linewidth]{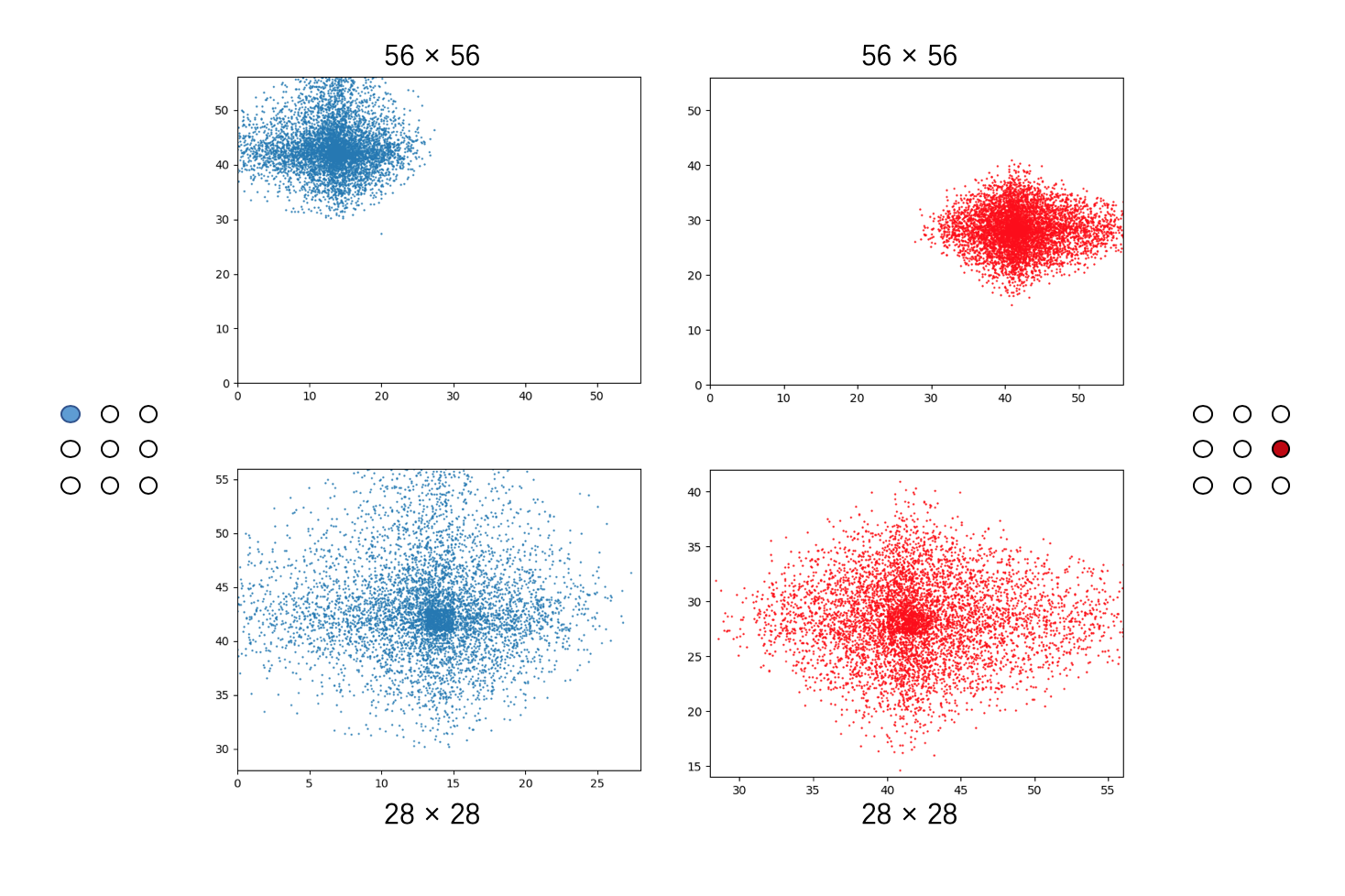} 
\caption{\textbf{Grid point specific representation region.} First row shows the distribution of ground truth location when the grid point heatmaps correspond to the whole proposal region. Blue point denotes top-left grid point, red point denotes mid-right grid point. Most of the pixels in supervision heatmap would never be activated. In Grid R-CNN Plus, the output heatmap is quartered from the original representation area according to it's corresponding grid point. Second row shows the new distribution of two kinds of grid points. We introduce random noise to make visualization better.}
\label{fig:gpsr}
\end{figure}

\subsection{Grid Point Specific Representation Region}
The most prominent update in Grid R-CNN Plus is the grid point specific representation. Since only positive samples ($IoU > 0.5$) will be selected to Grid branch, the ground truth label is constrained to a small region on the supervision map. As shown in figure~\ref{fig:gpsr}, the distribution of ground truth label is closely related to the category of its corresponding grid point. In a typical $3 \times 3$ grid points case, the ground truth labels of top-left grid point can only appear in the upper left region of the whole supervision heatmap. In fact, if all grid points of an object share the same representation area, for each grid point, most of its output should never be activated, which is very inefficient.

To solve the problem above, we propose grid point specific representation. Specifically, we reduce the output size of grid branch from original $56 \times 56$ to $28 \times 28$. For each grid point, the new output represents its most probable quarter region of the original area. Using grid point specific representation, we can halve the output size without changing the representation resolution. As illustrated in figure \ref{fig:gpsr}, grid point specific representation can also be approximated as a normalization process which shift a biased distribution to a normalized one.

\begin{table*}[!h]
\begin{center}
\begin{tabular}{ l | c | c | c  c | c  c  c | c }
\hline
method & backbone & AP & $\text{AP}_{.5}$ & $\text{AP}_{.75}$ & $\text{AP}_{S}$ & $\text{AP}_{M}$ & $\text{AP}_{L}$ & inference speed \\
\hline
Faster R-CNN w FPN~\cite{lin2017feature} & ResNet-50 & 37.4 & 59.3 & 40.3 & 21.8 & 40.9 & 47.9 & 0.09 \\
Grid R-CNN w FPN~\cite{Lu_2019_CVPR} & ResNet-50 & 39.6 & 58.3 & 42.4 & 22.6 & 43.8 & 51.5 & 0.25 \\
Grid R-CNN Plus w FPN & ResNet-50 & \textbf{40.4} & 58.6 & 43.7 & 23.2 & 44.2 & 52.4 & 0.11 \\
\hline
Faster R-CNN w FPN~\cite{lin2017feature} & ResNet-101 & 39.5 & 61.2 & 43.1 & 22.7 & 43.7 & 50.8 & 0.12 \\
Grid R-CNN w FPN~\cite{Lu_2019_CVPR} & ResNet-101 & 41.3 & 60.3 & 44.4 & 23.4 & 45.8 & 54.1 & 0.29 \\
Grid R-CNN Plus w FPN & ResNet-101 & \textbf{42.0} & 60.6 & 45.4 & 24.1 & 46.2 & 55.2 & 0.13 \\
\hline
\end{tabular}
\caption{Bounding box detection AP on COCO \textit{minival}. Grid R-CNN Plus outperforms the original one on both speed and accuracy.}
\label{tab:coco}
\end{center}
\end{table*}

\subsection{Light Grid Head }
Since the output size of the grid branch is halved, as described above, we simultaneously reduce the resolution of all other features in grid branch(i.e. from $14 \times 14$ to $7 \times 7$). This update helps to significantly reduce the overall computation cost of the grid branch. Specifically, in grid branch, features of each proposal are extracted by RoIAlign operation with fixed spatial size of $14 \times 14$, followed by one $3 \times 3$ stride 2 conv layer. After that, seven $3 \times 3$ stride 1 conv layers are adopted to generate $7 \times 7$ resolution features. Then we split the feature into N groups(default as 9), each corresponds to a grid point. Since we explicitly assign these split features to grid points, output heatmaps with $28 \times 28$ resolution can be generated by two 2x \textbf{group} deconvolutional layers. The up-sampling process is fast due to grouping operation.

Another good property obtained from grid point specific representation is that features from different grid points are normalized closer. It means we don't need that much convolution layers to cover the gap between grid points. In updated model, we use one $5 \times 5$ depth-wise convolution layer instead of original three consecutive convolution layers in feature fusion module to achieve comparable performance.

\subsection{Image-across Sampling Strategy}
Since the grid branch only uses positive proposals for training, the varying numbers of positive samples in different sampled batches have a non-negligible impact on the performance. For example, some images may have only a few positive samples while others may have hundreds of positive samples. In this case, the feature distribution in grid branch could be unstable. In Grid R-CNN Plus, we utilize an image-across sampling strategy. The positive proposals are sampled across several images. When few positive samples can be sampled from one image, other images can provide more to fill the vacancy. Specifically, during training, we sample at most 192 positive proposals across 2 images instead of sampling at most 96 proposals per image. This strategy makes the training more robust and improves the performance.

\subsection{Non-Maximum Suppression Only Once}
In original Grid R-CNN, proposals are first classified by classification branch. Then non-maximum suppression with 0.5 IoU threshold is adopted to remove duplicate samples according to their classification score. Only top 125 high score samples will be selected and processed by Grid branch for further localization. After that another NMS operation will be adopted to generate the final results. According to our experiments, even with a small set of proposals, running NMS along 80 categories (COCO dataset) is still very slow. So in Grid R-CNN Plus, we remove the second NMS for acceleration. In first NMS, the IoU threshold is set to 0.3 and classification score threshold is set to 0.03, only top 100 proposals will be selected.

\section{Experiments}
We conduct experiments on ResNet-50 and ResNet-101 backbone with FPN~\cite{lin2017feature} baseline. We train our model on the union of 80k train images and 35k subset of val images and test on a 5k subset of val(minival. Results in table\ref{tab:coco} show that Grid R-CNN Plus runs faster and performs better than previous Grid R-CNN.

Different from original paper, we use group normalization~\cite{wu2018group} instead of sync batch normalization. The number of groups should be set to multiple of the number of grid points. In typical $3 \times 3$ grid points configuration, we set the number of groups to 36 in conv layers and 9 in deconv layers. Note that group normalization~\cite{wu2018group} is only used in grid branch. 

Other training and testing configurations are same as original paper~\cite{Lu_2019_CVPR} unless described above. Inference time test is done on TITANXp GPU in our GPU cluster. We also reproduce Grid R-CNN based on mmdetection~\cite{mmdetection2018}, an open source object detection framework. Code is available at \href{https://github.com/STVIR/Grid-R-CNN}{https://github.com/STVIR/Grid-R-CNN}.

{\small
\bibliographystyle{ieee_fullname}
\bibliography{egbib}

\begin{thebibliography}{1}\itemsep=-1pt

\bibitem{mmdetection2018}
Kai Chen, Jiangmiao Pang, Jiaqi Wang, Yu Xiong, Xiaoxiao Li, Shuyang Sun,
  Wansen Feng, Ziwei Liu, Jianping Shi, Wanli Ouyang, Chen~Change Loy, and
  Dahua Lin.
\newblock mmdetection.
\newblock \url{https://github.com/open-mmlab/mmdetection}, 2018.

\bibitem{girshick2015fast}
Ross Girshick.
\newblock Fast r-cnn.
\newblock In {\em Proceedings of the IEEE international conference on computer
  vision}, pages 1440--1448, 2015.

\bibitem{he2017mask}
Kaiming He, Georgia Gkioxari, Piotr Doll{\'a}r, and Ross Girshick.
\newblock Mask r-cnn.
\newblock In {\em Computer Vision (ICCV), 2017 IEEE International Conference
  on}, pages 2980--2988. IEEE, 2017.

\bibitem{he2016deep}
Kaiming He, Xiangyu Zhang, Shaoqing Ren, and Jian Sun.
\newblock Deep residual learning for image recognition.
\newblock In {\em Proceedings of the IEEE conference on computer vision and
  pattern recognition}, pages 770--778, 2016.

\bibitem{lin2017feature}
Tsung-Yi Lin, Piotr Doll{\'a}r, Ross Girshick, Kaiming He, Bharath Hariharan,
  and Serge Belongie.
\newblock Feature pyramid networks for object detection.
\newblock In {\em Computer Vision and Pattern Recognition (CVPR), 2017 IEEE
  Conference on}, pages 936--944. IEEE, 2017.

\bibitem{long2015fully}
Jonathan Long, Evan Shelhamer, and Trevor Darrell.
\newblock Fully convolutional networks for semantic segmentation.
\newblock In {\em Proceedings of the IEEE conference on computer vision and
  pattern recognition}, pages 3431--3440, 2015.

\bibitem{Lu_2019_CVPR}
Xin Lu, Buyu Li, Yuxin Yue, Quanquan Li, and Junjie Yan.
\newblock Grid r-cnn.
\newblock In {\em The IEEE Conference on Computer Vision and Pattern
  Recognition (CVPR)}, June 2019.

\bibitem{ren2015faster}
Shaoqing Ren, Kaiming He, Ross Girshick, and Jian Sun.
\newblock Faster r-cnn: Towards real-time object detection with region proposal
  networks.
\newblock In {\em Advances in neural information processing systems}, pages
  91--99, 2015.

\bibitem{wu2018group}
Yuxin Wu and Kaiming He.
\newblock Group normalization.
\newblock In {\em Proceedings of the European Conference on Computer Vision
  (ECCV)}, pages 3--19, 2018.

\end{thebibliography}
}

\end{document}